\documentclass[journal]{IEEEtran}

\usepackage{stfloats}
\usepackage{mathrsfs}
\usepackage{graphicx}
\usepackage{amsmath}
\usepackage{natbib}
\usepackage{lipsum}

\ifCLASSINFOpdf
\else
\fi
\begin{document}
%
\title{Graph-based Semantical Extractive Text Analysis}
%
%
%

\author{Mina Samizadeh
\thanks{Mina Samizadeh is with the Department
of Computer and Information Sciences, University of Delaware. Email: minasmz@udel.edu}
}

%
%

\markboth{Journal of \LaTeX\ Class Files,~Vol.~6, No.~1, January~2007}%
{Shell \MakeLowercase{\textit{et al.}}: Bare Demo of IEEEtran.cls for Journals}
%



\maketitle
\thispagestyle{empty}

\begin{abstract}
In the past few decades, there has been an explosion in the amount of available data  produced from various sources with different topics. The availability of this enormous data necessitates us to adopt effective computational tools to explore the data. This leads to an intense growing interest in the research community to develop computational methods focused on processing this text data. A line of study focused on condensing the text so that we are able to get a higher level of understanding in a shorter time. The two important tasks to do this are keyword extraction and text summarization. In keyword extraction, we are interested in finding the key important words from a text. This makes us familiar with the general topic of a text. In text summarization, we are interested in producing a short-length text which includes important information about the document. The TextRank algorithm, an unsupervised learning method that is an extension of the PageRank (algorithm which is the base algorithm of Google search engine for searching pages and ranking them) has shown its efficacy in large-scale text mining, especially for text summarization and keyword extraction. this algorithm can automatically extract the important parts of a text (keywords or sentences) and declare them as the result. However, this algorithm neglects the semantic similarity between the different parts. In this work, we improved the results of the TextRank algorithm by incorporating the semantic similarity between parts of the text. Aside from keyword extraction and text summarization, we develop a topic clustering algorithm based on our framework which can be used individually or as a part of generating the summary to overcome coverage problems.
\end{abstract}


%
\IEEEpeerreviewmaketitle

\section{Introduction}
%
%
%
%

The goal of text summarization is extracting a few important sentences from the document while preserving the main idea of the text. A good summary keeps the main topic of the text simultaneously, occupy less space than the original document. This is a very complex problem because it needs to emulate the cognitive capacity of human beings to generate summaries and still is an open problem in natural language processing. Since it is a difficult task most of the researches of literature focused on the extractive aspect of summarization which returns important sentences of documents without any change in the sentences in contrast to the abstractive aspect which generates new sentences that reveal the main topic of the text and it requires more resources to be trained. A summary generated by an automatic text summarizer should consist of the most relevant information in a document and at the same time, it should be condensed to take less space in comparison to the source document. Nevertheless, automatically generating summaries is a challenging task. 
An issue in extractive text summarization is the amount of summary coverage and diversity of topics in the input document. In an optimized summarizer, results cover a sufficient amount of topics especially in situations which the document contains multiple topics.

In an attempt to solve these issues, in this work we proposed an extractive summarization method based on the TextRank algorithm to find important sentences in a documents and using a word2vec model to include semantics in the summary.
Moreover, we proposed a topic clustering method which can also be used as a post-processing technique on the text summarization results. In this topic-clustring method, we consider semantic as well as similar keywords to measure the similarities between sentences in a text and categorize them together. Further, we proposed a keyword extraction method based on the textRank algorithm which, similar to the proposed text summarization method, considers the semantics as well as the statistics of the words in the sentences.


To summarize our works:
\begin{itemize}
    \item Proposing a graph-based extractive semantical text summarization to summarize texts in any language
    \item Proposing a graph-based extractive semantical keyword extraction (n-gram extraction)
    \item Proposing a semantical topic clustering method
    \item Validating the performance of proposed methods on real-world English and Persian datasets. 
\end{itemize}
In the following section, we provide a brief description of the word embeddings and text summarization methods. The proposed methods are described in Section III, followed by experimental results in Section IV. Final remarks and a discussion about our plans are reported in Section V.

\section{Related Works}

\subsection{Word Embeddings}

Word embedding is a term using to describe a set of language modeling and feature learning techniques in natural language processing (NLP) where words or phrases are mapped to real-valued vectors. It involves a mathematical process where words from one dimension space embed into a higher dimensional continuous vector space, which is often tens to hundreds. A word embedding is a trained representation for text where words with similar semantic have a similar representation. In the word embedding process, each individual word is represented by a real-valued vector with dimensionally regard to trained vector space. Since, in these techniques, each word is demonstrated by a vector which is learned through a neural network training process, word embedding is often referred to as a deep learning method. The stepping stone idea to this method is using a densely distributed representation for each word. In these methods, each word represents by tens or hundreds of dimensions, contrary to sparse word representations, such as one-hot encoding which often has thousands or millions of dimensions. In \cite{bengio2003neural} \cite{schwenk2007continuous} \cite{mikolov2010recurrent} models each individual word is transformed into a real-valued vector using a pre-trained lookup table. The neural network language model can be used to obtain word representation. Which can further be utilized in other tasks, for example, \cite{collobert2008unified} \cite{turian2010word} embedding used for classification in NLP task, or using feed-forward networks \cite{bengio2003neural} \cite{gauvain2006neural} and then recurrent neural network models \cite{mikolov2010recurrent} \cite{mikolov2011extensions} to predict the probability distribution of the next word. Many other methods have been proposed in order to create word embeddings that are based on the Distributed Hypothesis \cite{harris1954distributional}. Among these methods, word2vec \cite{mikolov2013linguistic} and Glove \cite{pennington2014glove} are the most popular methods with roughly the same accuracy. Word2vec consists of two models, continuous bag-of-words (CBOW) and skip-gram. These models learn a vector representation of each word by using a neural network language model and can be trained efficiently on the large size of the corpus. Word2vec allows models to learn the complex semantic relationship between words by using vector operations. For instance, Equation \ref{eq:1} represents that if we subtract the embedded vector of Italy from Rome (Subtract the country from its capital), then add France (another country), we get a vector approximately similar to the vector of Paris (capital of France). This example shows that the model learned the semantics of text and captured that the relationship between Italy and Rome is similar to the relationship between France and Paris.

\begin{equation} \label{eq:1}
 \vec{v}(Rome)- \vec{v}(Italy)+ \vec{v}(France)\approx \vec{v}(Paris)
\end{equation}
  
In this work, we have adopted the implementation of Doc2vec from Gensim Python library. The goal of Doc2vec is to create a vector representation of a document, regardless of its length. But documents do not have logical structures such as words, so another method has to be found. While a word vector is trying to represent the concepts of the word, a document vector purpose is to represent the concept of a document. Mikolov and Le \cite{le2014distributed}  employed an easy and inventive concept. They used Word2vec model and added another vector (Paragraph ID) to it to overcome the length difficulty. It has two different models, PV-DM (Distributed Memory version of Paragraph Vector) and PV-DBOW (Distributed Bag of Words of Paragraph Vectors), the former is a type of extension to CBOW model in Word2vec, but instead of just using words to predict the next word, it uses another feature vector which is unique per document, as well. It plays a role as a memory that remembers what is missing from the current context. The later is more similar to the skip-gram model in Word2vec and uses a distributed bag of words. In Gensim implementation there is a variable (dm) which correlates to selecting one of these models. By default, it is 1 which is PV-DM and it is used in our proposed algorithm.

\subsection{Text Summarization}
The first summarization method proposed by \cite{luhn1958automatic}, tried to weight the sentences of a document based on the frequency of words and omitting the very high-frequency common words. Since then the research community has widely addressed automatic text summarization techniques. Very good surveys are available and have proposed by many researchers  \cite{sehgal2018modification} \cite{pal2014approach} \cite{saranyamol2014survey} \cite{gambhir2017recent}, but since we have exploited a graph-based method, we describe graph-based summarization and then explain the TextRank algorithm specifically which is the ground algorithm of our proposed method.

In graph-based methods, every sentence is assumed as a node of a graph and two sentences are connected with an edge if they possessed some words in common, or in other words, there is an edge between two sentences if with a similarity measure (such as cosine) their similarity were more than a threshold. In this regard, we can conclude two things from the generated graph. First, the isolated partitions in the graph (sub-graphs that are isolated from the rest of the sub-graphs) that embodies a distinct topic in the text. In query-specific summaries, some sentences from each sub-graphs may return as an answer to that specific question, while for common summaries, most important and representative sentences may be chosen from each of the sub-graphs, or they can be assumed as different topics in the original text and the summary for covering all the topics, should select sentences from all of the sub-graphs. The second inference can be obtained by graph-theoretic methods to find the most important and representative sentences (nodes) in the graph. This can be done by assuming more important sentences have a higher degree (more nodes are connected to them) thus, they have a higher probability to be included in the summary. By assuming documents as a graph which nodes are sentences and edges are the similarity between the nodes many different kinds of graph theory tools and measurement have been applied to find the important sentences for extractive summarization work \cite{mihalcea2005language}.

Now we want to explain the TextRank algorithm \cite{mihalcea2004graph} which is a very high-performance extractive summarization technique and its foundation is the PageRank algorithm. PageRank algorithm is commonly used in the Google search engine to compute the rank of web pages. This algorithm works with these main insights: Important pages are linked by important pages and the PageRank value of a page is essentially the probability of a user visiting that page. Scores in the PageRank algorithm calculated as follow in \ref{eq:2}:

\begin{equation}
P_{r}(V_{i})=(1-d)+d*\sum_{V_{j}\in ln(V_{i})} \frac{P_{r}(V_{j})}{|Out(V_{j})|}
\label{eq:2}
\end{equation}

In the TextRank, sentences are considered equivalent to web pages and apply the PageRank algorithm over sentences in the graph. Formally, let $G=(V, E)$ be a directed graph with the set of vertices $V$ and set of edges $E$, where $E$ is a subset of $V \times V$. For a given vertex $V_{i}$, let $In(V_{i})$ be the set of vertices that point to it (predecessors), and let $Out(V_{i})$ be the set of vertices that vertex $V_{i}$ points to (successors). The score of a vertex $V_{i}$ is defined as Equation \ref{eq:3} in \cite{page1999pagerank}:

\begin{equation}
S(V_{i})=(1-d)+d*\sum_{V_{j}\in ln(V_{i})} \frac{1}{|Out(V_{j})|}S(V_{j})
\label{eq:3}
\end{equation}

Where $d$ is a damping factor and it is a number between $0$ and $1$ and has the role of compounding the probability of going from a specific vertex to another random vertex in the graph, into the model. In the web surfing concept, the "random surf model" is adopted in this graph-based ranking algorithm. In this model, when a user clicks on links at random with a probability of d, it goes to a new page with the probability $1-d$. By starting from an arbitrary value assigned to each node in the graph, the computation iterates until it reaches a convergence point below a given threshold. After running the algorithm, each vertex is associated with a score, representing the “importance” of the vertex within the graph \cite{mihalcea2004graph}.

\section{PROPOSED METHOD}

We have proposed a graph-based extractive text summarization algorithm which is based on the TextRank algorithm. The essence of the TextRank algorithm is vertex voting, where the voting equals an edge between two vertexes. If a specific vertex has higher dependency and similarity to the rest of the vertexes, it would be mapped with a higher value. For illustrating this consider $G=(V, E)$ as a graph with $V$ being the vertex set and $E$ the edge set. The importance metric of each vertex is as shown in Equation \ref{eq:4}:

\begin{equation}
W(V_{i})=(1-d)+d*\sum_{j\in ln(V_{i})} \frac{W_{ji}}{\sum _{V_{k} \in Out(v_{j}) W_{jk}}}W(V_{j})
\label{eq:4}
\end{equation}

\begin{figure}
    \centering
    \includegraphics[width=4cm]{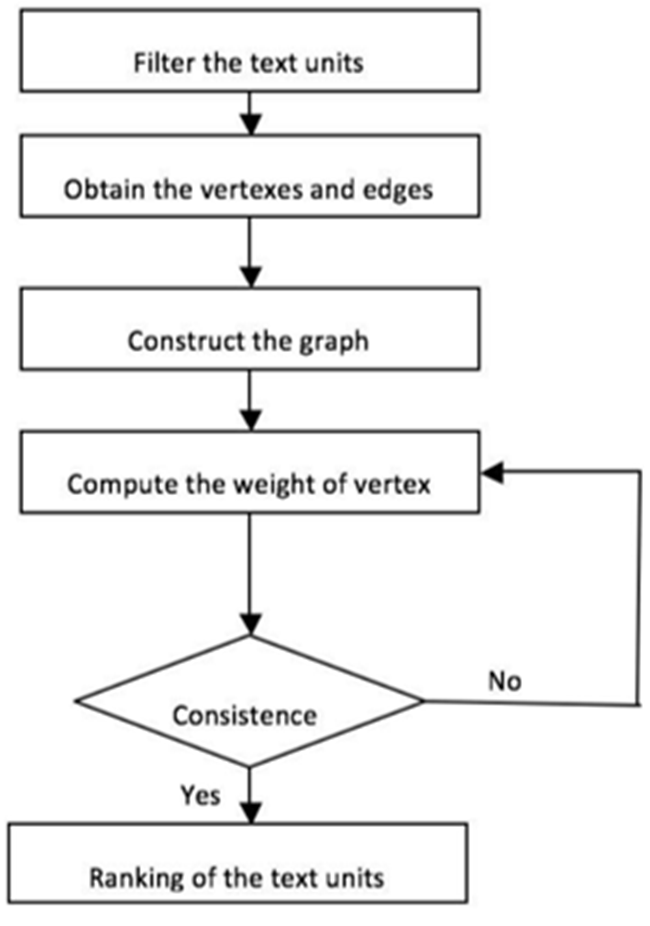}
    \caption{TextRank Algorithm Flowchart}
    \label{fig:1}
\end{figure}

$In(V_{i})$ is the set of indexes for vertexes (text units which are sentences here) and have a common window with $V_{i}$ in linear order in a sentence, $Out(V_{j})$ is the set of vertexes that have a common window with $V_{i}$, $d$ is a damping factor, its default value is 0.85. Commonly, a weight assigned to an edge from $V_{j}$ to $V_{i}$ as $w_{ji}$ and it is computed by calculating the chances of occurrence of two text units in the same text window with a fixed size, with the ordinary size of 2. In the initialization weight of all the text units are equal to one and all of the weights obtain consistency after some iteration through Equation (4). The text units (sentences) which have more weight are considered the key text units (sentences). Figure \ref{fig:1} demonstrates the flowchart of the algorithm.

In our work, we have used TextRank as a voting method to vote among doc2vec scores of each sentence. In other words, instead of using number of similar words in each sentence as similarity measure between sentences (nodes in graph) as it is in the TextRank, our algorithm scores the relationship between the sentences based on the score get from doc2vec trained model; hence, the semantics of words and sentences are included in the similarity rather than just considering the same words to measure the similarity. In the above flowchart, we made some alternations in the phase of computing the weight of the vertex. The novelty of our method is including semantic into the TextRank summarization and just not focus on a statistical analysis of similar exact words. Besides, it has a significant advantage, unsupervised learning, of the TextRank algorithm which makes it needless to huge corpus for training. This gives the algorithm the facility to be very convenient to utilize for summarizing new text, efficiently.

For implementing this algorithm we have used TextRank from Gensim, a Python library, and altered some part of it to become compatible with the Persian language to test on Persian corpus. We have trained two Doc2vec models, one on Hamshahri corpus \cite{aleahmad2009hamshahri} for Persian language usage, and another on Text8 corpus (which is a pre-processed
a version of the first 100 million characters from Wikipedia dump) for English language purpose. Then we have added them to the algorithm to weigh the edges of the graph (similarity score between sentences). We could yield the most important sentences in the graph (nodes with a higher score) with this implementation. By assuming that the most important sentences contain more information about the text and have more relations with other nodes; we could return them as the summarization of the input text. In this implementation which is available at Github \footnote{https://github.com/minasmz/Persian-Summarization}, we can choose the number of words to be included in the summarization, as well as the ratio of input text we want to get as summarization, which is 20 percent of text as a default. To overcome the coverage problem in multi-topic texts we proposed a clustering method and apply the semantic text summarization described above. The proposed clustering algorithm weighs the similarity between paragraphs in a row based on a doc2vec model trained on separate documents. If the score was above a threshold (mean + standard deviation) they would merge in the same cluster; otherwise, they would not and we assume them as one cluster by itself. Then we apply the text summarization algorithm on each cluster independently. In this way, we can cover all topics of input text in our summarization.

\subsection{topic clustering}
In some scenarios, when the size of the text which want to summarize becomes large, the coverage issue arise. In the coverage issue, the text summarizer returns the most frequent sentences as the summary. However, the text may content other important sentences which might be neglected due to a lower frequency. To overcome this issue, we added a topic clustering method before applying the text summarization method on our text data. To do so, we apply the doc2vec  method on the paragraphs  of training set and obtained the representation of paragraphs of training set. Then, the distance between the paragraphs in training set is measured using the cosine similarity. Further, we calculate the mean $(m)$ and standard deviation of distance $(std)$ in the training set. To place two paragraphs in a same cluster of topic, we assumed they should have a distance less than the $m + std$, otherwise they belong to two different topics cluster. When we obtain the cluster of topics, we apply the text summarization method described earlier on each clusters. In this way, we resolve the coverage problem in text summarization. 

\subsection{keyword extraction}
After clustering the topics and summarizing the text, we tried to extract the keywords in the summaries.
Two different methods are introduced to extract the n-grams as keywords in a text. In the first methods, the semantic of the words are not included in the algorithm and just the frequency of words using tf-idf method is used. In this method we have adopted the bm25 scoring function which regularly is used as a keyword extractor. We applied some alternation to the available bm25 algorithm in Gensim library of Python and made it compatible with Persian language. In the second method, the semantic of the words in the text is considered as well. To include the semantic of words, we trained a word2vec model and took a similar method introduced in text summarization to generate a graph of words. In this graph, each word is a node of the graph and the weight of edges are the cosine similarity between words representations connecting to each other. Then by applying the TextRank algorithm on the generated graph, we obtain the most important words. In this approach not only frequency of words are included in the obtaining the keywords, but also the meanings of them are included. In both methods, we tried to return n-gram (by default 10-gram) as the keyword. After acquiring most important and frequent words, we check them in unigram to n-gram of the words $(n=10)$ if frequency of n-gram be more than half of the frequency of the word and its occurrence be more than 2 (in big size inputs this number should be increased) the important word occurred in the text would reduce to the bigger n-gram. Finally we return the most important n-gram where the important words are in them.

\section{Experimental Results}

\begin{figure}[htp]
    \centering
    \includegraphics[width=8cm]{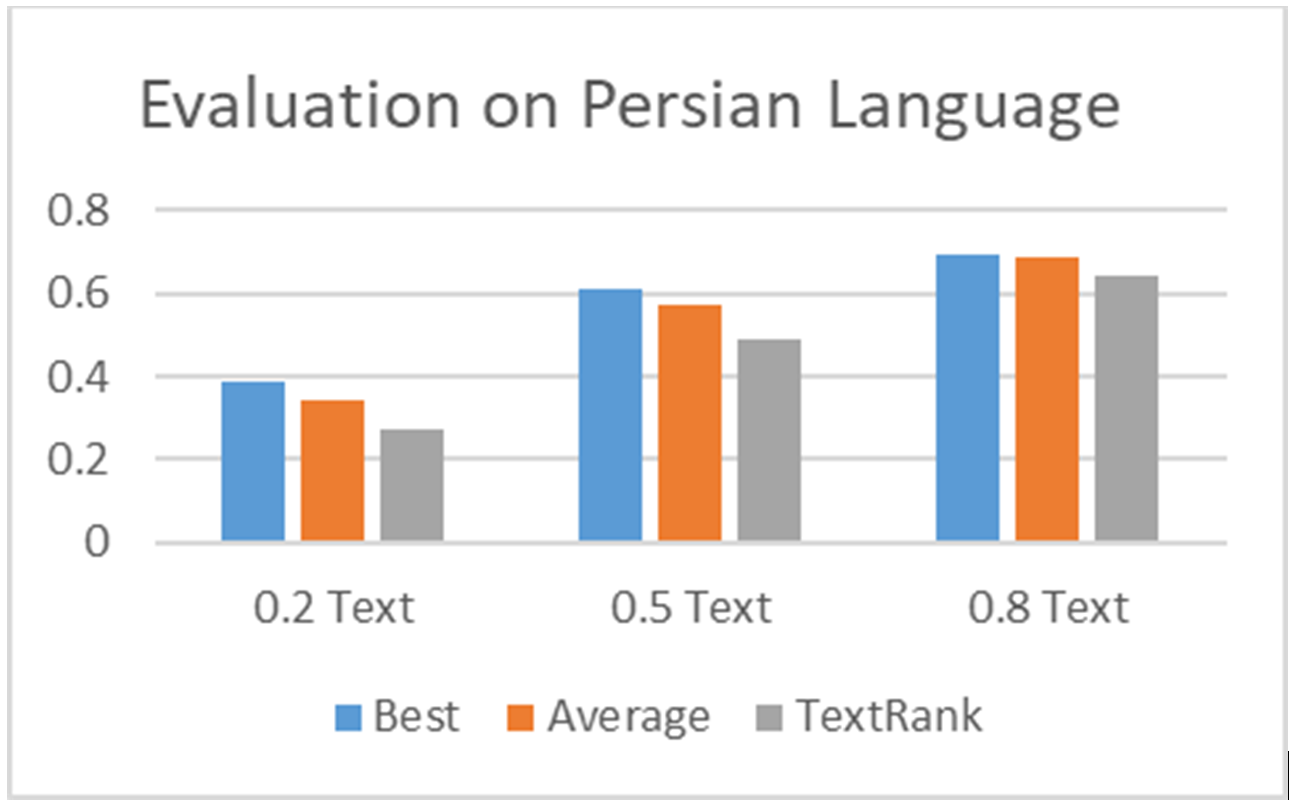}
    \caption{Evaluation on Persian Language}
    \label{fig:2}
\end{figure}

\begin{figure}[htp]
    \centering
    \includegraphics[width=8cm]{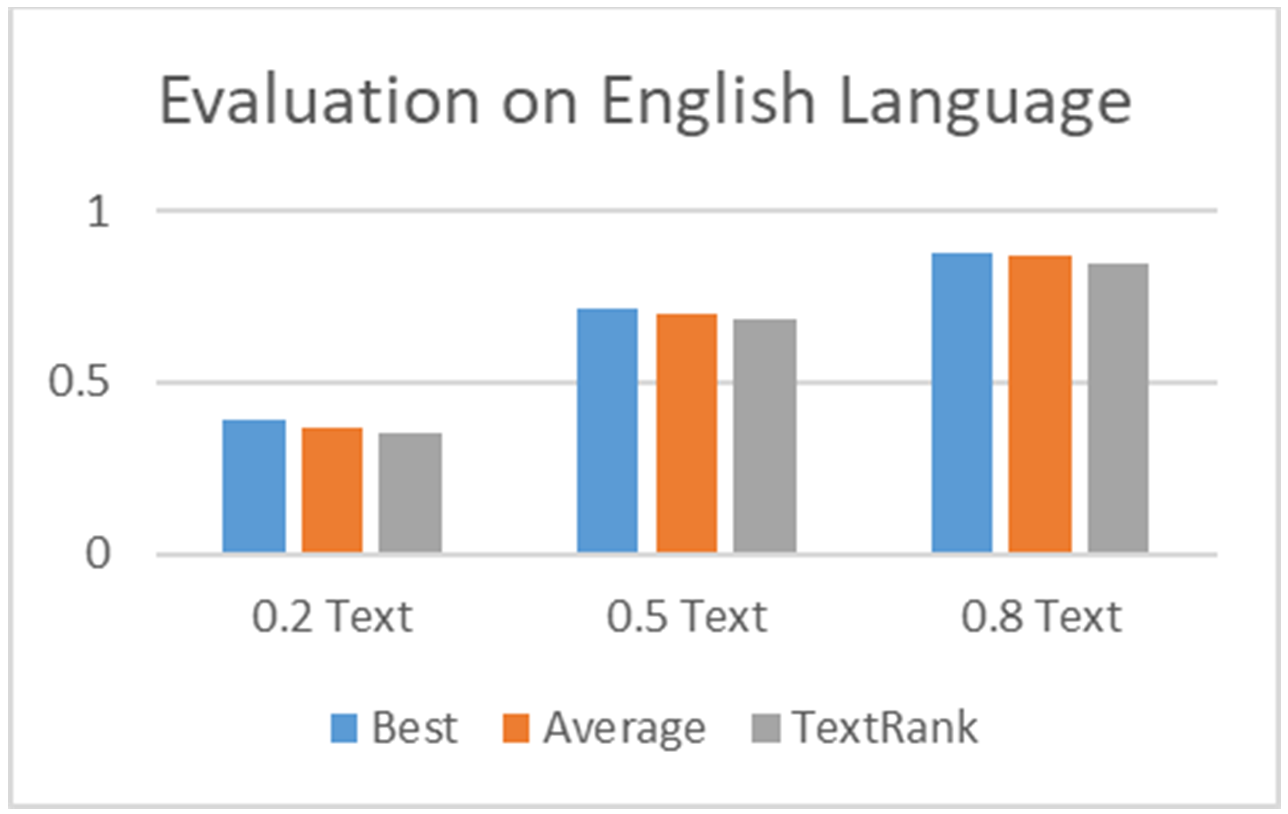}
    \caption{Evaluation on English Language}
    \label{fig:3}
\end{figure}

Although in text summarization there is no unique way to summarize a text and it is a difficult task even for the human to reach a concise form of a text, but researchers prepared some gold test set for evaluating the methods on them. We have used Zamanifar dataset for testing the algorithm on the Persian language and a BBC News \cite{greene06icml} dataset which contains BBC news and their summaries for evaluating the work in the English language. We have calculated the accuracy by Rough measurement Equation \ref{eq:5}. It should be mentioned that doc2vec involves some randomness because of the negative sampling that is being used in its implementation, a different set of negative examples would be tried in each call so the result of the inferred vector from doc2vec model would be different in each call. To overcome this and measure the accuracy of the algorithm precisely, we have run the algorithm 10 times on each text and calculated the average of these 10 results. We compare the result of each two datasets separately in Persian and English languages to the pure TextRank algorithm. We have also set the ratio of summaries to 0.2, 0.5, and 0.8, respectively; and compare the results on each language which has been illustrated in Figure(\ref{fig:2}) and Figure(\ref{fig:3}).

\begin{align}  
& Rough\_2= \label{eq:5} \nonumber \\ 
& \dfrac{\small {\sum \limits_{s \in (RefSummaries)}\sum \limits_{bigrams_{i} \in S } \small{{min(count(i,x), count(i,S))}}}}{\sum \limits_{s \in \{RefSummaries\}} \sum \limits_{bigrams i \in S} count(i,S)} 
\end{align}

Results from the Persian language are illustrated in Figure(2). The best result is the highest rough score obtained from 10 times of running the algorithm on each text. And in the table you can see the average of the best score on 59 documents in Zamanifar dataset for the Persian language, the average score is averaged over 10 times running the algorithm, in the table averaging over average score in 59 document is brought and then it is compared to the result of the TextRank algorithm on 59 documents.

Figure(3), demonstrates the results of the algorithm on the English language, there is 417 document in the politic folder of BBC news dataset \cite{greene06icml} and the Rough measurement calculated on them, so the results are averaging over 417 documents.

The average of 10 runs on each document on average has comparable and better accuracy than TextRank The best summary among 10 times running the algorithm has the highest average on all of the experiments. This makes us conclude that the result of combining semantic in graph-based text summarization, TextRank, would be helpful to increase the performance and accuracy of the summaries in TextRank algorithm. Moreover, we can conclude that by having a better model of each language, higher accuracy would be obtained.

Due to the lack of gold standard set for keyword extraction and topic clustering, we experimentally observed that our methods genrate meaningful results.

\section{FUTURE WORKS AND DISCUSSION}

Text summarization is a difficult task which even humans could not reach a unique summary from a text. For decades, researchers in extractive text summarization tried to select the most important sentences from the text which contain useful information by weighing and ranking the sentences based on some statistical assumptions and neglect the semantics of the words in sentences. In the proposed method we can observe from the results that including semantic in summaries can improve the accuracy. In this work, we have employed doc2vec models for learning semantic. For future work, we can say that by having better preprocessed and more data relating to the context of input texts, we can train more accurate models which leads to better results. hence, in future works, we can train these models on a bigger and cleaner dataset for reaching better accuracy. Also, we can train and adopt other word embeddings to include semantic in the statistical algorithms. Moreover, In the current work, we have used TextRank since it performs better among other graph-based algorithms in text summarization. Also, we can inject the semantic of texts into the other available graph-based statistical text summarization algorithms, such as hit and evaluate the results of that algorithms, too.


\bibliographystyle{apalike}
\bibliography{journal}

\begin{thebibliography}{}

\bibitem[AleAhmad et~al., 2009]{aleahmad2009hamshahri}
AleAhmad, A., Amiri, H., Darrudi, E., Rahgozar, M., and Oroumchian, F. (2009).
\newblock Hamshahri: A standard persian text collection.
\newblock {\em Knowledge-Based Systems}, 22(5):382--387.

\bibitem[Bengio et~al., 2003]{bengio2003neural}
Bengio, Y., Ducharme, R., Vincent, P., and Jauvin, C. (2003).
\newblock A neural probabilistic language model.
\newblock {\em Journal of machine learning research}, 3(Feb):1137--1155.

\bibitem[Collobert and Weston, 2008]{collobert2008unified}
Collobert, R. and Weston, J. (2008).
\newblock A unified architecture for natural language processing: Deep neural
  networks with multitask learning.
\newblock In {\em Proceedings of the 25th international conference on Machine
  learning}, pages 160--167. ACM.

\bibitem[Gambhir and Gupta, 2017]{gambhir2017recent}
Gambhir, M. and Gupta, V. (2017).
\newblock Recent automatic text summarization techniques: a survey.
\newblock {\em Artificial Intelligence Review}, 47(1):1--66.

\bibitem[Gauvain et~al., 2006]{gauvain2006neural}
Gauvain, L., Bengio, Y., and Schwenk, H. (2006).
\newblock Neural probabilistic language models.
\newblock {\em Innovations in Machine}.

\bibitem[Greene and Cunningham, 2006]{greene06icml}
Greene, D. and Cunningham, P. (2006).
\newblock Practical solutions to the problem of diagonal dominance in kernel
  document clustering.
\newblock In {\em Proc. 23rd International Conference on Machine learning
  (ICML'06)}, pages 377--384. ACM Press.

\bibitem[Harris, 1954]{harris1954distributional}
Harris, Z. (1954).
\newblock Distributional structure. word, 10 (2-3): 146--162. reprinted in
  fodor, j. a and katz, jj (eds.), readings in the philosophy of language.

\bibitem[Le and Mikolov, 2014]{le2014distributed}
Le, Q. and Mikolov, T. (2014).
\newblock Distributed representations of sentences and documents.
\newblock In {\em International conference on machine learning}, pages
  1188--1196.

\bibitem[Luhn, 1958]{luhn1958automatic}
Luhn, H.~P. (1958).
\newblock The automatic creation of literature abstracts.
\newblock {\em IBM Journal of research and development}, 2(2):159--165.

\bibitem[Mihalcea, 2004]{mihalcea2004graph}
Mihalcea, R. (2004).
\newblock Graph-based ranking algorithms for sentence extraction, applied to
  text summarization.
\newblock In {\em Proceedings of the ACL Interactive Poster and Demonstration
  Sessions}, pages 170--173.

\bibitem[Mihalcea, 2005]{mihalcea2005language}
Mihalcea, R. (2005).
\newblock Language independent extractive summarization.
\newblock In {\em ACL}, volume~5, pages 49--52.

\bibitem[Mikolov et~al., 2010]{mikolov2010recurrent}
Mikolov, T., Karafi{\'a}t, M., Burget, L., {\v{C}}ernock{\`y}, J., and
  Khudanpur, S. (2010).
\newblock Recurrent neural network based language model.
\newblock In {\em Eleventh annual conference of the international speech
  communication association}.

\bibitem[Mikolov et~al., 2011]{mikolov2011extensions}
Mikolov, T., Kombrink, S., Burget, L., {\v{C}}ernock{\`y}, J., and Khudanpur,
  S. (2011).
\newblock Extensions of recurrent neural network language model.
\newblock In {\em 2011 IEEE International Conference on Acoustics, Speech and
  Signal Processing (ICASSP)}, pages 5528--5531. IEEE.

\bibitem[Mikolov et~al., 2013]{mikolov2013linguistic}
Mikolov, T., Yih, W.-t., and Zweig, G. (2013).
\newblock Linguistic regularities in continuous space word representations.
\newblock In {\em Proceedings of the 2013 Conference of the North American
  Chapter of the Association for Computational Linguistics: Human Language
  Technologies}, pages 746--751.

\bibitem[Page et~al., 1999]{page1999pagerank}
Page, L., Brin, S., Motwani, R., and Winograd, T. (1999).
\newblock The pagerank citation ranking: Bringing order to the web.
\newblock Technical report, Stanford InfoLab.

\bibitem[Pal and Saha, 2014]{pal2014approach}
Pal, A.~R. and Saha, D. (2014).
\newblock An approach to automatic text summarization using wordnet.
\newblock In {\em 2014 IEEE International Advance Computing Conference (IACC)},
  pages 1169--1173. IEEE.

\bibitem[Pennington et~al., 2014]{pennington2014glove}
Pennington, J., Socher, R., and Manning, C. (2014).
\newblock Glove: Global vectors for word representation.
\newblock In {\em Proceedings of the 2014 conference on empirical methods in
  natural language processing (EMNLP)}, pages 1532--1543.

\bibitem[Saranyamol and Sindhu, 2014]{saranyamol2014survey}
Saranyamol, C. and Sindhu, L. (2014).
\newblock A survey on automatic text summarization.
\newblock {\em Int. J. Comput. Sci. Inf. Technol}, 5(6):7889--7893.

\bibitem[Schwenk, 2007]{schwenk2007continuous}
Schwenk, H. (2007).
\newblock Continuous space language models.
\newblock {\em Computer Speech \& Language}, 21(3):492--518.

\bibitem[Sehgal et~al., 2018]{sehgal2018modification}
Sehgal, S., Kumar, B., Rampal, L., Chaliya, A., et~al. (2018).
\newblock A modification to graph based approach for extraction based automatic
  text summarization.
\newblock In {\em Progress in Advanced Computing and Intelligent Engineering},
  pages 373--378. Springer.

\bibitem[Turian et~al., 2010]{turian2010word}
Turian, J., Ratinov, L., and Bengio, Y. (2010).
\newblock Word representations: a simple and general method for semi-supervised
  learning.
\newblock In {\em Proceedings of the 48th annual meeting of the association for
  computational linguistics}, pages 384--394. Association for Computational
  Linguistics.

\end{thebibliography}


%

\ifCLASSOPTIONcaptionsoff
  \newpage
\fi

\end{document}